%% file: main.tex
\documentclass[pmlr,twocolumn,10pt]{jmlr} %

\usepackage{booktabs}
\usepackage{siunitx}

\usepackage[switch]{lineno}

\usepackage{tikz}
\usepackage{xspace}
\usepackage{titlesec}
\usepackage[normalem]{ulem}
\usepackage{caption}
\usepackage{etoolbox}

\usepackage{float}
\usepackage{listings}
\usepackage{xcolor}
\usepackage{multicol}

\definecolor{lightgray}{gray}{0.9}
\lstdefinestyle{promptstyle}{
    basicstyle=\ttfamily\scriptsize, %
    frame=single,
    breaklines=true,
    backgroundcolor=\color{lightgray}, %
    rulecolor=\color{black},
}
\lstdefinestyle{outputstyle}{
    basicstyle=\ttfamily\scriptsize, %
    frame=single,
    breaklines=true,
    backgroundcolor=\color{white}, %
    rulecolor=\color{black},
}

\let\oldcitep\citep
\renewcommand{\citep}[2][]{\begingroup\small\oldcitep[#1]{#2}\endgroup}

\let\oldcite\cite
\renewcommand{\cite}[2][]{\begingroup\small\oldcite[#1]{#2}\endgroup}

\newtheorem{assumption}{Assumption}
\theorembodyfont{\upshape}
\theoremheaderfont{\scshape}
\theorempostheader{:}
\theoremsep{\newline}

\jmlrvolume{259}
\jmlryear{2024}
\jmlrsubmitted{LEAVE UNSET}
\jmlrpublished{LEAVE UNSET}
\jmlrworkshop{Machine Learning for Health (ML4H) 2024} %

 \title[Learning Explainable Treatment Policies with
Clinician-Informed Representations]{Learning Explainable Treatment Policies with
Clinician-Informed Representations: A Practical
Approach}

\author{%
\Name{Johannes O. Ferstad} \\ %
\addr OpenAI (work done while at Stanford University)
\AND
\Name{Emily B.~Fox}\\ %
\addr Stanford University \& CZ Biohub
\AND
\Name{David Scheinker} \\ %
\Name{Ramesh Johari} \Email{rjohari@stanford.edu}\\
\addr Stanford University
}

\begin{document}

\maketitle

\begin{abstract}
Digital health interventions (DHIs) and remote patient monitoring (RPM) have shown great potential in improving chronic disease management through personalized care. However, barriers like limited efficacy and workload concerns hinder adoption of existing DHIs, and limited sample sizes and lack of interpretability limit the effectiveness and adoption of purely black-box algorithmic DHIs.  In this paper, we address these challenges by developing a pipeline for learning explainable treatment policies for RPM-enabled DHIs.

We apply our approach in the real-world setting of RPM using a DHI to improve glycemic control of youth with type 1 diabetes.  Our main contribution is to reveal the importance of {\em clinical domain knowledge} in developing state and action representations for effective, efficient, and interpretable targeting policies. We observe that policies learned from clinician-informed representations are significantly more efficacious and efficient than policies learned from black-box representations.

This work emphasizes the importance of collaboration between ML researchers and clinicians for developing effective DHIs in the real world.
\end{abstract}

\paragraph*{Data and Code Availability}
We use data from three IRB-approved clinical trials of remote monitoring of $N = 281$  patients with type 1 diabetes (T1D) \citep{scheinker_new_2022, prahalad_teamwork_2022, prahalad_equitable_2024}.  These data are not yet publicly available.  The code used to generate our results is available at \href{http://github.com/jferstad/ml4h-explainable-policies}{http://github.com/jferstad/ml4h-explainable-policies}.

\paragraph*{Institutional Review Board (IRB)}
Our research is carried out under approval by the Stanford University Institutional Review Board, under protocol 52812.

\vspace{-0.6\baselineskip}
\section{Introduction}
\label{sec:intro}
\vspace{-0.4\baselineskip}

Digital health interventions (DHIs) and remote patient monitoring (RPM) have the potential to revolutionize patient care with treatment strategies dynamically personalized to each patient's characteristics and context. DHIs and RPM have been associated with improved management of many of chronic conditions including heart disease, diabetes, and mental health \citep{prahalad_equitable_2024, whitelaw_barriers_2021, liverpool_engaging_2020}. Relative to standards of care that rely on fixed-cadence clinic visits, RPM-enabled DHIs promise more timely, personalized, and frequent patient support \citep{scheinker_new_2022}. These technologies could help move population-level outcomes towards those typically seen in environments with more healthcare resources, particularly for underserved communities \citep{tikkanen_multinational_2017, anderson_its_2003, rodriguez_racial_2017, prahalad_equitable_2024}.  

We consider RPM-enabled DHIs that involve the following typical workflow.  On a regular cadence (e.g., weekly), the RPM platform takes as input rich, high-dimensional patient data, including granular data from sensors such as continuous glucose monitors (CGMs) and activity trackers. These data are used to form a representation of {\em patient states}, based on which patients are prioritized to receive a DHI (or {\em action}).  Actions may include messaging the patient, recommending activity or treatment adjustments (e.g., a dose change). A {\em targeting policy} determines how to rank patients for interventions based on patient state. The results of this ranking inform a whole-population care model in which clinicians determine what actions to take (e.g., which patients to message and what to say in the message).

While the majority of clinicians plan to use RPM-enabled DHIs, their adoption has been limited by significant, well-documented challenges \citep{stevens_effectiveness_2022, peterson_digital_2024, vivalink_remote_2023, lawrence_operational_2023, sasangohar_remote_2018, cresswell_organizational_2013, borges_do_nascimento_barriers_2023, borghouts_barriers_2021}. {\em First} is uncertain efficacy, i.e., the difficulty of learning effective policies. A major hurdle here is the difficulty of developing representations of patient states from high-dimensional patient-level data from relatively few patients (e.g., at most a few hundred).  {\em Second} is workload concerns: because clinical teams have limited capacity, the targeting policy must respect resource constraints by directing attention to patients who will benefit most from intervention. {\em Third} is a difficulty understanding or interpreting the technology.  This lack of interpretability leads to a reluctance of care teams to adopt solutions that rely on black-box models.

In this paper we develop a pipeline to support the optimization of a care model that addresses the preceding concerns.  Our work is carried out in the real-world context of an RPM-enabled DHI for individuals with type 1 diabetes (T1D).  In the setting we consider, as in \citet{prahalad_equitable_2024}, patients wear a continuous glucose monitor (CGM) that measures glucose every 5 minutes, generating a high-dimensional time series. The real-world deployment in \citet{prahalad_equitable_2024} generates patient states using clinically-informed single-dimensional summaries of this CGM data, such as the percentage of time glucose levels are in a desired target range (70-180 mg/dL).  Clinicians provide guidance on how to improve management through telehealth interactions using natural language messages sent to the patient (e.g., ``Increase your pre-dinner insulin dose") based on the dashboard presentation of recent patient CGM data.  The objective of technology-based RPM for T1D is to improve patients' glucose management on an ongoing basis, through personalized, timely interventions.

Our main contribution is to reveal the importance of {\em clinical domain knowledge} in developing state and action representations for effective, efficient, and interpretable targeting policies.  Because real-world settings have limited sample sizes, the inductive bias from clinical domain knowledge provides significant benefits to real-world policy performance.  In particular, in the preceding T1D RPM context, we evaluate several approaches to low-dimensional state and action representations: from black-box machine learning methods, to clinician-informed learned representations.
{\em We observe that policies derived from clinician-informed representations significantly outperform policies learned from black-box-learned representations in terms of efficacy and efficiency.}  In fact, our evaluation reveals that learned policies outperform random targeting only when the state and action representations are clinically informed -- amplifying the importance of clinical inductive bias in practice.  Further, the use of clinical domain knowledge also yields policies that are more interpretable than black-box policies with state and action representations that maintain clinically relevant features and interventions.

To carry out our evaluation, we develop an end-to-end pipeline for learning targeting policies: we (1) learn low-dimensional state and action representations; (2) construct targeting policies by ranking patients based on estimated conditional average treatment effects (CATEs); and (3) evaluate the policies in the presence of capacity constraints.  %
While each component of this pipeline has been studied in practice, our paper presents a coherent implementation of these steps together to carry out the evaluation described above.  This pipeline may be of independent interest to digital health researchers carrying out similar optimization and evaluation of targeting policies in other real-world settings.  

Our approach is broadly applicable to the evaluation and optimization of digital health interventions. Notably, our work suggests that interaction between machine learning researchers and healthcare domain experts is essential for developing practical, effective, and interpretable data-driven treatment policies that can be adopted in clinical practice.

\vspace{-1\baselineskip}
\section{Related work}
\vspace{-0.4\baselineskip}

Many studies have focused on the offline evaluation of conditional average treatment effect (CATE) estimators and of learned treatment policies, which are foundational to our approach \citep{mahajan_empirical_2023, dwivedi_stable_2020, tang_model_2021, yadlowsky_evaluating_2021, sverdrup_qini_2023, bouneffouf_survey_2020, feuerriegel_causal_2024, imai_experimental_2023, imai_statistical_2024}. Our work combines methods from these works into a novel pipeline. Like us, they seek to facilitate the development of better treatment policies. But unlike their work, we do not treat our state and action representations as fixed. Instead, we learn and evaluate targeting policies across many different representations, including interpretable representations defined by clinicians.

Our focus on explainable and interpretable causal inference and machine learning methods in healthcare aligns with the growing recognition of the importance of interpretability in this domain. \citet{rasheed_explainable_2022} provide a survey of explainable machine learning methods in healthcare, highlighting the need for transparent and understandable models. \citet{mello_denialartificial_2024} discuss the practical challenges associated with non-interpretable recommendations generated by machine learning models broadly used by insurance companies. Our approach complements this line of research by demonstrating the value of clinician-informed representations for learning interpretable treatment policies in digital health interventions.

\vspace{-1\baselineskip}
\section{Data and context: States, actions, rewards}
\label{sec:context}
\vspace{-0.4\baselineskip}

We use data from three IRB-approved clinical trials of remote monitoring of $N = 281$  patients with type 1 diabetes (T1D) \citep{scheinker_new_2022, prahalad_teamwork_2022, prahalad_equitable_2024}. Participants in the trials wear CGMs that regularly transmit glucose measurements to TIDE, a remote patient monitoring platform \citep{ferstad_population-level_2021, kim2024adaptation}. At regular intervals, e.g., weekly, clinicians use TIDE to review patient CGM data and decide whether to send treatment recommendations through asynchronous secure messaging (Figure \ref{fig:rpm}). Due to constraints on provider time, at each review interval TIDE presents data for a subset of patients prioritized based on simple metrics from the consensus guidelines established by the American Diabetes Association, e.g., patients with a relatively high percentage of very low CGM readings.  

Our data consist of variable numbers of days of data for each patient, depending on when the patient started the study; we let $T_i$ denote the number of days of data for patient $i$.  We let $X_{it}^d$ denote patient demographics for patient $i$ on day $t$; $X_{it}^d$ is a combination of time-invariant patient demographics like sex and race/ethnicity, and time-variant demographics like age and insulin pump use.  In addition, clinicians and the TIDE dashboard consider the previous two weeks of CGM readings---taken every 5 minutes--in determining patient prioritization for intervention; we let $X_{it}^g \in \Re^{4032}$ to be the (high-dimensional) vector of CGM recordings for patient $i$ over the two previous weeks prior to day $t$ (which may include missing values due to, e.g., the patient not wearing their CGM).  As such, the individual $X_{it}^g$ are defined using a day-by-day rolling window on the raw CGM trace for patient $i$.  Taken together, we call $X_{it} = (X_{it}^d, X_{it}^g)$ the {\em high-dimensional patient state} for patient $i$ at time $t$; we let $\mathcal{X}$ denote the state space.

Given our clinical context, we focus on messages as the action taken by clinicians.  In particular,  we let $M_{it}$ denote the raw text of treatment messages sent to patient $i$ on day $t$.  If no message is sent, then $M_{it} = 0$.  We refer to $M_{it}$ as the {\em high-dimensional action} taken on patient $i$ at time $t$; we let $\mathcal{M}$ denote the action space.  Note that in our data, patients are rarely messaged more than once per week. 

At a high level, the goal of any targeting policy is to direct clinicians' limited resources to take actions (i.e., send appropriate messages) to those patients who are most in need of intervention (given their patient state).  Such a policy is considered effective if it leads to improvements in patients' glucose management.  In particular, we define the {\em reward} $r_{it}$ to be the improvement in the {\em time-in-range} (TIR) of patient $i$ at day $t$ in the subsequent week relative to the prior week.  
TIR is the fraction of a patient's glucose readings between 70-180 mg/dL, one of the most commonly used outcome metrics in T1D care \citep{battelino_clinical_2019}.

\vspace{-0.5\baselineskip}
\section{Methods: A pipeline for policy learning and evaluation}
\label{sec:methods}
\vspace{-0.4\baselineskip}

\begin{figure*}[t]
  \centering
  \begin{minipage}[t]{0.26\linewidth}
    \centering
    \includegraphics[width=\textwidth]{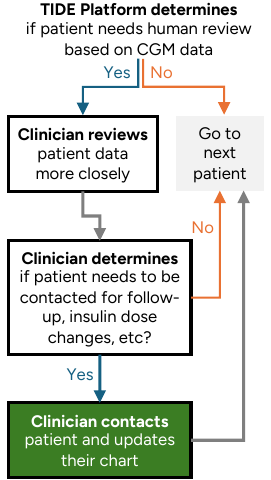}
    \captionsetup{format=plain, width=\linewidth}
    \caption{Remote patient monitoring workflow.}
    \label{fig:rpm}
  \end{minipage}
  \hfill
  \begin{minipage}[t]{0.67\linewidth}
    \centering
    \includegraphics[width=\textwidth]{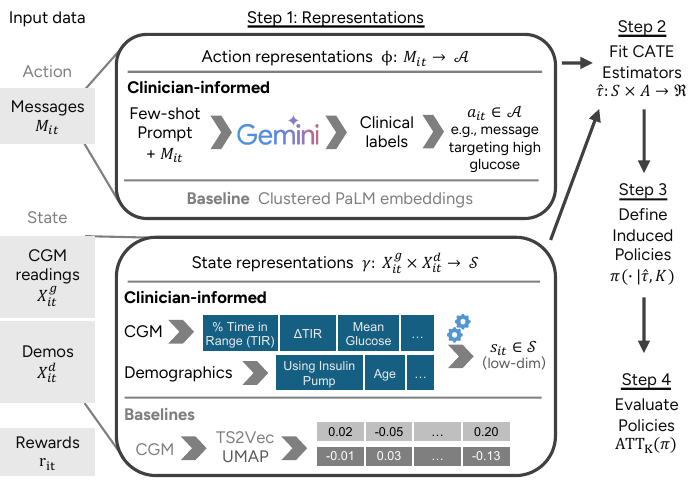}
    \caption{Diagram of pipeline for learning targeting policies.}
    \label{fig:data_flows}
  \end{minipage}
  \vspace{-0.7\baselineskip}
\end{figure*}

In this section, we outline the three step approach to learning targeting policies for remote patient monitoring of T1D: (1) learning low-dimensional state and action representations; (2) constructing targeting policies by ranking patients based on estimated conditional average treatment effects (CATEs); and (3) evaluating the policies in the presence of capacity constraints (Figure \ref{fig:data_flows}).  Our approach is applicable to other domains with similar characteristics; where possible we describe each step using general notation and specialize as appropriate to our specific clinical context.

\vspace{-0.6\baselineskip}
\subsection{State and action representations}
\label{sec:state_action_reps}
\vspace{-0.4\baselineskip}

In our setting both $X_{it}$ and $M_{it}$ are high-dimensional relative to the number of patients.  For this reason, we require dimension-reduced representations of both states and actions.  We let $\mathcal{S}$ (resp., $\mathcal{A}$) denote the space of low-dimensional states (resp., actions). Formally, we let $\gamma: \mathcal{X} \rightarrow \mathcal{S}$ be a function that maps each patient's high-dimensional state to a low-dimensional state representation $s_{it} \in \mathcal{S}$, i.e., $s_{it} = \gamma(X_{it})$.  Similarly, we let $\phi: \mathcal{M} \rightarrow \mathcal{A}$ be a function that maps the high-dimensional action $M_{it}$ to an action representation $a_{it} \in \mathcal{A}$, i.e., $a_{it} = \phi(M_{it})$.  We assume that $0 \in \mathcal{A}$, and that $\phi(0) = 0$ uniquely (i.e., the control action maps to itself, and is the only action to do so).  For simplicity, we also assume the set $\mathcal{A}$ is {\em finite} in our development.

We simplify by assuming that patient states are drawn from a stationary {\em superpopulation} distribution $\mathbb{P}$.  Of course, in practice, there may be complex temporal dynamics in the patient state; in general, we presume that any such dynamics can be captured through the high-dimensional underlying state representation $X_{it}$ (e.g., the rich CGM sensor data capturing the trajectory of the patient).
With this superpopulation view, we let $R(x,m)$ denote the reward obtained by a patient in state $x$ if they receive action $m$; we can view $R(x,m)$ as the potential outcomes for a patient in state $x$, if we vary the action $m$ \citep{robins_estimation_1994}.  Note that then $r_{it} = R(X_{it}, M_{it})$, and that given an action $m$, the quantity $\mathbb{E}[R(x,m) | \gamma(x) = s]$ is the expectation of reward (over the superpopulation distribution of patient states) with respect to the dimension-reduced representation.  Throughout the paper, we make the following {\em conditional consistency} assumption; this assumption is similar in nature to the consistency assumption on outcomes in causal inference \citep{hernan_does_2016}, except that we have adapted it to apply to the action representation selected.  

\begin{assumption}
\label{assumption:cond_consistency}
{\em Conditional consistency of action representation.} 
An action representation $\phi$ is {\em conditionally consistent} if for all $s \in \mathcal{S}$ and $m,m' \in \mathcal{M}$ such that $\phi(m) = \phi(m')$, there holds $\mathbb{E}[R(x,m) | \gamma(x) = s] = \mathbb{E}[R(x,m') | \gamma(x) = s]$.
\end{assumption}

If $\phi$ satisfies Assumption \ref{assumption:cond_consistency}, then for any action $a \in \mathcal{A}$, the quantity $\mathbb{E}[R(x,m) | \gamma(x) = s]$ is the same for any $m$ such that $\phi(m) = a$; thus we can define the reward in terms of the dimension-reduced representations as $\rho(s,a) = \mathbb{E}[R(x,m) | \gamma(x) = s]$ where $m$ satisfies $\phi(m) = a$.

We use two approaches to constructing representations: (1) algorithmic black-box approaches, and (2) clinician-informed representations.  

{\em Black-box baselines: Low-dimensional representations directly from raw data}.  For action representation, we get features from the raw message text by generating 728-dimensional embeddings using PaLM (Pathways Language Model) \citep{anil_palm_2023}, a large-scale autoregressive language model. Then we cluster the embeddings into discrete message types using K-means to define discrete actions.
For state representations, we consider two methods for learning low-dimensional state representations directly from the raw CGM traces: TS2Vec \citep{yue_ts2vec_2021}, a universal representation learning framework for time series that applies hierarchical contrastive learning; and UMAP (Uniform Manifold Approximation and Projection), a non-linear dimensionality reduction technique that preserves the local and global structure of the data \citep{mcinnes_umap_2018}.

{\em Clinician-informed representations.}  In most clinical contexts with high-dimensional states or actions (e.g., sensor time series, imaging, text, etc.), clinicians already have a lower-dimensional set of features they extract for clinical decision-making.  Rather than starting from the raw data representation, our approach 
learns a low-dimensional representation starting from this ``medium''-dimensional feature representation.  The inductive bias provided by such domain knowledge will prove crucial to learning performant, interpretable policies that are clinically grounded.

 For action representations, we extract interpretable clinical features using few-shot labeling with Gemini Pro \citep{gemini_gemini_2024} to generate labels from each message (e.g., was a dose change recommended, did the dose change focus on high or low glucose). %
We use the features most predictive of rewards to anchor a set of discrete actions; we then group each of the remaining features with its closest anchor action, or to an ``other message type'' category, based on the similarity of their clinical meanings.
See Appendix \ref{appendix:actionrep} for details. 

For state representations, we %
start with a ``medium''-dimensional set of pre-defined CGM clinical features %
commonly used in practice: time-in-range (TIR; 70-180 mg/dl), mean glucose, time below 70 mg/dl, time below 55 mg/dl, etc. \citep{battelino_clinical_2019}. We also include demographics (e.g., age, language preference, insurance type, insulin pump use, etc.).  See Appendix \ref{appendix:staterep} for a full list of included clinician-defined state features.  We evaluate representations that are different subsets of these features: the full set, a learned subset predictive of rewards, and a subset defined most relevant by clinicians.

\vspace{-0.6\baselineskip}
\subsection{Targeting policies: Ranking via estimated CATEs}
\vspace{-0.4\baselineskip}

A {\em targeting policy} chooses which patients to prioritize for treatment, and what actions to choose for them, given a capacity constraint.  We consider targeting policies that rank patients according to CATEs estimated given the dimension-reduced state and action representations.

A CATE function estimates the effect of an action conditional on the patient's state. The {\em true} CATE function, denoted $\tau: \mathcal{S} \times \mathcal{A} \rightarrow \mathbb{R}$, is $\tau(s, a) = \rho(s,a) - \rho(s,0)$.  (Note $\tau(s,0) = 0$ for the control action.)  Given a dataset $\mathcal{D} = \{(X_{it}, M_{it}, r_{it})\}_{i,t}$, an {\em estimated CATE function} $\hat{\tau}: S \times A \rightarrow \mathbb{R}$ is a learned estimate of the expected treatment effect for a given state and action representation (defined by $\gamma$ and $\phi$, respectively).

We estimate CATE functions $\hat{\tau}$ using several estimators. The S-Learner (Single Learner) is a simple approach that trains a single model to predict the outcome using both the action and state representations as input features. The T-Learner (Two Learners) trains separate models for each treatment action and for a pre-defined control action (e.g., no message), and then estimates the CATE as the difference between their predictions. The X-Learner (X-Learner) \citep{kunzel_meta-learners_2017} is a meta-learner that estimates the CATE by training separate models for each action, and then training a final model on the imputed treatment effects relative to the control action. The Causal Forest \citep{wager_estimation_2015} is an extension of random forests that estimates the CATE by recursively partitioning the data based on the covariates and treatment assignment. The DR Forest (Doubly Robust Forest) \citep{athey_generalized_2019} is a variant of the Causal Forest that combines the estimates from a propensity and outcome model to achieve double robustness. We also create an ensemble estimator by combining all of the previous estimators, where the final CATE estimate is obtained by taking a weighted average of the predictions of the individual estimators. The weights for each model in the ensemble are learned using the validation dataset \citep{mahajan_empirical_2023}. We fit the CATE estimators using EconML \citep{syrgkanis_causal_2021} and predictions from nuisance models (prediction propensities and outcomes) trained with AutoML in FLAML \citep{wang_flaml_2021}.

For an estimated CATE function $\hat{\tau}$ and a capacity constraint $K \leq N$ on the number of patients that can receive a treatment $a \neq 0$ (i.e., other than the control treatment), we define a targeting policy that prioritizes treating the patients with the highest estimated rewards under the optimal actions.

\begin{definition}[Induced targeting policy]\label{def:induced_policy}
Fix an estimated CATE function $\hat{\tau}$, capacity constraint $K \leq N$, and a patient state vector $s = (s_1, \ldots, s_N)$.  The targeting policy $\pi = \pi(\cdot | \hat{\tau}, K)$ induced by $\hat{\tau}$ and $K$ chooses actions for each patient as follows:

\noindent 1. For each patient $i$, let $a_i^* = \arg\max_{\tilde{a} \in \mathcal{A}} \hat{\tau}(s_i, \tilde{a})$.\\
2. Rank the patients in descending order of $\hat{\tau}(s_i, a_i^*)$.\\
3. For the top $K$ ranked patients, set $\pi_i(s) = a_i^*$; for the remaining patients set $\pi_i(s) = 0$.
\end{definition}

\vspace{-0.6\baselineskip}
\subsection{Evaluating targeting policies}
\vspace{-0.4\baselineskip}

Our goal is to find combinations of state representation $\gamma$, action representation $\phi$, and CATE function $\hat{\tau}$ such that the resulting induced targeting policy $\pi(\cdot | \hat{\tau}, K)$ achieves high quality outcomes, i.e., actually targets patients with the highest treatment effects.

Formally, for any targeting policy $\pi$ with capacity constraint $K$, we define the value function $\mathsf{ATT}_K(\pi)$ as the {\em average treatment effect on the treated (ATT)}:
\( \mathsf{ATT}_K(\pi) = \mathbb{E}\left[ \frac{1}{K} \sum_{i = 1}^N \tau(s, \pi_i(s)) \right] \).
Here the expectation is over the superpopulation, where we assume that patients are sampled i.i.d.~from the superpopulation.  Note that for any patient that receives the control action under $\pi$, the treatment effect in the sum is zero.  As a result, if the policy $\pi$ only provides non-control actions to at most $K$ patients, the right hand side will be the average treatment effect of the treated patients.

Our goal is to learn policies with high ATT, given the capacity constraint $K$.  Further, in practice, we will be interested in additional qualitative desiderata, e.g., whether the resulting policy is interpretable or aligns with clinical guidelines.  In our empirical evaluation we will test whether these requirements are met, and in particular, whether the use of clinician-informed representations biases selected policies towards being interpretable as well.  

Before continuing we comment briefly on the {\em optimal} policy.  In particular, for fixed $K$, let $\pi^* = \pi(\cdot | \tau, K)$; this is the policy that ranks patients according to their true treatment effects.  It is easy to check that $\mathsf{ATT}_K(\pi^*) \geq \mathsf{ATT}_K(\pi)$ for any other policy $\pi$ that targets at most $K$ patients; see Proposition \ref{prop:pi_star} in Appendix \ref{appendix:proofs} for a proof.  We show in the following theorem that if we estimate $\tau$ effectively, then the value of the estimated optimal policy converges to the value of the true optimal policy; the result is analogous to existing results for policy learning \citep{wager_estimation_2015}.  See Appendix \ref{appendix:proofs} for proof details.

\begin{theorem}
\label{thm:optimality}
For each $N$, let $\hat{\tau}_N$ be a CATE estimator such that $\sup_{s \in \mathcal{S},a \in \mathcal{A}} | \tau(s,a) - \hat{\tau}_N(s,a) | \to 0$ in distribution.   Suppose also that the treatment effects are bounded: $\sup_{s \in \mathcal{S},a \in \mathcal{A}} |\tau(s,a)| < \infty$.  Consider a sequence $K_N$ such that $K_N/N \to c$ as $N \to \infty$, with $0 \leq c \leq 1$.  Let $\pi_N = \pi(\cdot | \hat{\tau}_N, K_N)$ be the associated sequence of targeting policies, and let $\pi_N^*$ be the associated sequence of optimal targeting policies.  Then $\mathsf{ATT}_{K_N}(\pi_N) - \mathsf{ATT}_{K_N}(\pi_N^*) \to 0$ as $N \to \infty$.
\end{theorem}

In general, for a policy $\pi$, to estimate $\mathsf{ATT}_K(\pi)$ for a given $K$, we require an estimate of the treatment effect of each action $a$, for each patient-day $(i,t)$ pair in our evaluation data.  A challenge here is that in our data, there is confounding between the actions and rewards. For example, clinicians are more likely to contact a patient with a recent drop in glucose control, and that patient is also more likely to have improved glucose control in the following week even if they are not contacted by a clinician (regression to the mean). If we fail to account for this confounding, we would overestimate the impact of interventions on the reward.  

To account for confounding, we use a doubly robust approach.  In particular, we adjust for a set of \emph{control covariates} $X_{it}^c = \bigl\{ \gamma^c(X_{it}^g), X_{i}^{d'} \bigl\}$. This is the representation of the patient state that clinicians see when reviewing patients, which includes a low-dimensional projection of the CGM data $\gamma^c(X_{it}^g)$ and a subset of the demographics $X_{it}^{d'} \subset X_{it}^d$.  We make the following (commonly used) assumptions to adjust for confounding and perform doubly robust policy evaluation.

\begin{assumption}
\label{assumption:consistency_sutva}
{\em Consistency and stable unit treatment value.}  The potential outcomes \citep{imbens_causal_2015} for each patient $i$ at time $t$ under treatment $M_{it} = m$ are the same as the observed outcomes if they actually received treatment $m$, and these potential outcomes depend only on the treatment $M_{it}$ assigned to that patient, not on the treatments assigned to other patients. Formally, $r_{it}(m) = r_{it} \hspace{0.5em} \text{if} \hspace{0.5em} M_{it} = m$.
\end{assumption}

\begin{assumption}
\label{assumption:condig}
{\em Conditional ignorability.}  Given the control covariates for patient $i$ at time $t$, $X_{it}^c$, the observed actions (treatment messages) $M_{it}$ are independent of the potential reward $r_{it}(m)$ for all possible actions $m \in \mathcal{M}$. Formally,
\begin{equation*}
r_{it}(m) \perp M_{it} \mid X_{it}^c, \quad \forall m \in \mathcal{M},
\end{equation*}
where $\mathcal{M}$ is the set of all possible actions (messages), and $r_{it}(m)$ denotes the potential reward for patient $i$ at time $t$ under action $m$.
\end{assumption}

Under these assumptions, we fit outcome models $\hat{r}(X^c, a)$ predicting the rewards under each action conditional on a vector of control covariates $X^c$, and a model predicting the reward under the control action $\hat{r}(X^c, 0)$.  We also fit models $\hat{e}(X^c, a)$, which estimates the propensity scores (probabilities of each action conditional on the control covariates). All of the nuisance models are trained with AutoML in FLAML \citep{wang_flaml_2021}.

Now suppose that a patient $i$ is observed at day $t$ in an evaluation dataset $\mathcal{E}$.  We define the following doubly robust score for each action $a$; this is an estimate of the treatment effect of action $a$ for patient $i$ at day $t$:
\label{eval:drscores}
\vspace{-0.5\baselineskip}
\begin{align*}
\hat{\Gamma}_{it}(a) &= (\hat{r}(X_{it}^c, a) - \hat{r}(X_{it}^c, 0)) \\
& + (r_{it} - \hat{r}(X_{it}^c, a_{it})) 
\left(\frac{I_{a_{it} = a}}{\hat{e}(X_{it}^c, a)} - \frac{I_{a_{it} = 0}}{\hat{e}(X_{it}^c, 0)}\right)
\end{align*}

The doubly robust score corrects for unmeasured confounding in the following way: if at most one of the reward model $\hat{r}$ or the propensity model $\hat{e}$ is misspecified, the doubly robust score will still be consistent for the true treatment effect.  In practice, we cannot completely rule out the possible simultaneous misspecification of both models.  However, as noted above, we take advantage of control covariates that capture all information available to clinicians at the time of choosing an action, helping mitigate bias due to confounding.

We estimate $\mathsf{ATT}_K(\pi)$ on an evaluation dataset $\mathcal{E}$ with $N$ patients, $K$ of whom are treated with a message, over $T$ time periods as:

\( \widehat{\mathsf{ATT}}_K(\pi) = \frac{1}{T} \sum_{t = 1}^T \frac{1}{K} \sum_{i = 1}^N \hat{\Gamma}_{it}(\pi_i(s_t)) \), where $s_t = (s_{1t}, \ldots, s_{Nt})$ is the patient state vector on day $t$. To estimate representative performance in a real clinic, we specifically report $\widehat{\mathsf{ATT}}_K(\pi)$ for $K/N = 0.25$; we refer to this metric as \emph{ATT@25\%}.  This metric estimates the performance of the policy when treating a proportion of the population that is of similar size to the capacity of our clinic.  

\paragraph*{Data splitting.} We split the data into three parts of approximately equal size by randomly dividing the {\em patients} into three groups and putting the data from each group of patients into separate datasets: train, validation, and test. The training data are used for pre-training state and action representations, CATE estimation, and to train nuisance models (if needed to adjust for confounding). The validation data are used to evaluate the performance of candidate representations and targeting policies induced by the CATE estimators.  The test data are used for valid estimation of the performance of the final chosen policy. By creating these splits across patients, we ensure that we learn a treatment policy that will generalize across patients from a similar population. 

\vspace{-1\baselineskip}
\section{Results}\label{sec:results}
\vspace{-0.4\baselineskip}

\textbf{Interpretability and clinical relevance of representations.} We expect to see correlation between states and actions if the representations capture how clinicians make decisions. In managing T1D, clinicians are concerned with {\em highs} (glucose levels above 180 mg/dL) and {\em lows} (glucose levels below 70 mg/dL).  When these events occur, we expect to see clinicians send messages targeting those events.

\begin{figure*}[t]
\centering 
\includegraphics[width=0.9\linewidth]{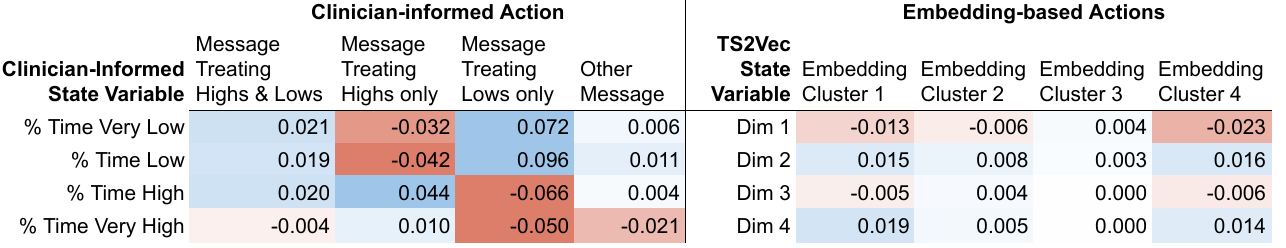} 
\caption{Pearson correlations between continuous state variables and binary action indicators.}
\label{fig:interpreting_reps} 
\end{figure*}

Figure \ref{fig:interpreting_reps} shows the correlations between continuous state variables and binary action indicators with clinician-informed (left) or black-box-learned (right) state and action representations. We see that clinician-informed state variables are correlated with clinician-informed actions: messages that target highs (resp., lows) are sent when states representing highs (resp., lows) are observed.  TS2Vec-learned state variables are less correlated with embedding-based actions.  In other words, the black-box-learned representations are not learning relationships that are relevant to clinicians' actions, while (as expected) the clinician-informed state and action representations are.

\begin{figure*}[t]
\centering 
\vspace{-1\baselineskip}
\includegraphics[width=1\linewidth]{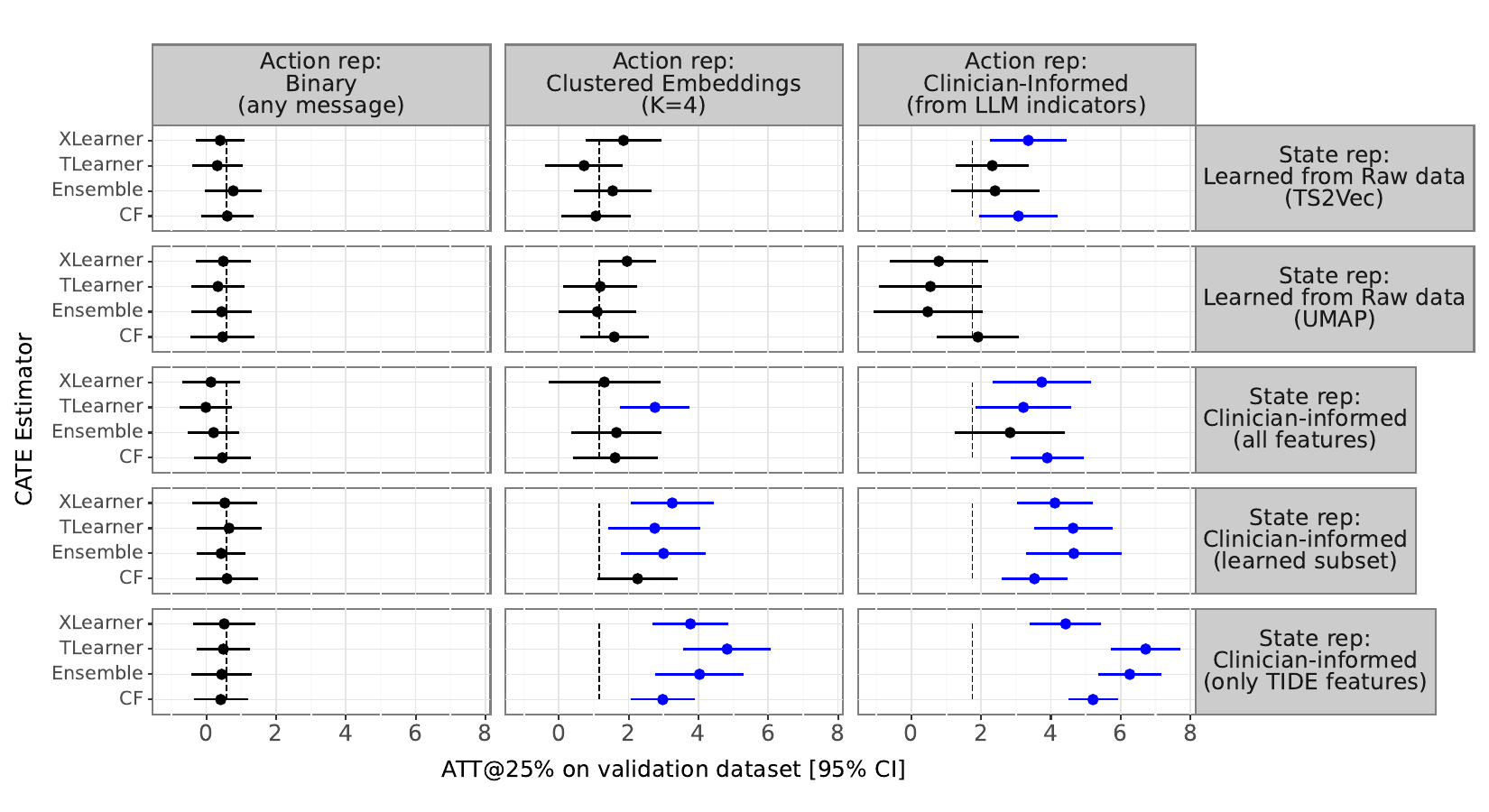} 
\vspace{-1.5\baselineskip}
\caption{Estimated ATT@25\% with 95\% CIs on validation data across representations and CATE estimators. ATEs of the action with the highest predicted ATE shown as vertical dashed lines (expected ATT under random targeting). See Appendix \ref{appendix:additional_results} for more results.}
\label{fig:att25} 
\end{figure*}

\textbf{Policy performance.}  Figure \ref{fig:att25} compares the ATT@25\% of different combinations of state and action representations across CATE estimators. We find that the estimated policy performance is equivalent to random targeting (ATE) for policies learned from most of the baseline representations we tested. Notably, policies derived from clinician-informed representations significantly outperform policies learned from black-box algorithmic baseline representations; not only are they more interpretable and clinically grounded, they also have higher efficacy.

A closer look at the state and action representations provides additional insight. For state representations, we see increasing performance with increasing levels of domain knowledge, moving from the ``medium''-dimensional representation (all clinician-informed features), to the learned subset, to the fully clinician-informed subset (TIDE features).  In this clinical setting, the TIDE-only features represent strong clinical domain knowledge of CGM features relevant to patient care, and have been developed over many years.  

By contrast, our clinician-informed action representations require a learning procedure, since we started with high-dimensional (unlabeled) text messages as our raw actions.  We again see the benefits of clinical inductive bias: the clinically-informed action set significantly outperforms a black-box-learned clustering. %
See Appendix \ref{appendix:additional_results} for additional policy evaluation results (including TOC curves).

After identifying the best-performing policy on the validation set, we evaluate it on the held-out test set to check for potential selection bias inflating our results on the validation data. The ATT@25\% for the policy (clinician-informed action representation, TIDE state representations, T-Learner) is 6.6 [95\% CI: 5.6-7.6] on the test set, which is similar to the validation set result of 6.7 [5.7-7.7].

\begin{figure*}[!t]
\centering 
\includegraphics[width=0.9\linewidth]{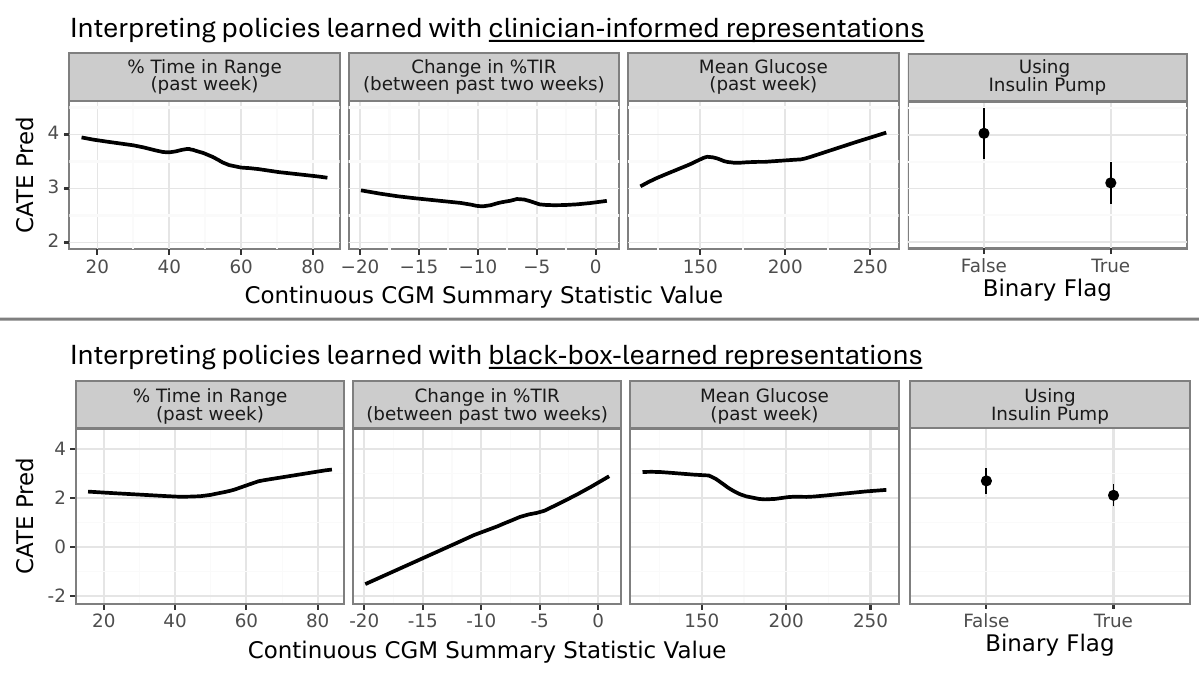} 
\vspace{-0.3\baselineskip}
\caption{CATE predictions for the optimal action for policies learned using both approaches. Policies learned from clinician-informed representations (top) are interpretable and align with clinical guidelines, while those learned from black-box-learned representations (bottom) do not.}
\label{fig:interpret} 
\end{figure*} 

\textbf{Policy interpretation and clinical alignment.}  
By inspecting how the CATE predictions vary across clinician-defined features we can assess if they align with clinical knowledge.  We want to recommend only those policies that align with clinical best practices \citep{battelino_clinical_2019}, since clinicians are unlikely to adopt the policy otherwise. 

Although management of T1D requires careful attention to both highs and lows, highs are much more consequential events for TIR than lows \cite{addala2021clinically}.  Lows are often emergent events requiring acute intervention, and also are coupled to other interventions (e.g., alarms on CGMs).  By contrast, highs are more persistent, significant, and longer-term in their impacts on TIR.  Since TIR is our reward, we expect that clinically aligned policies should focus primarily on reducing highs \cite{addala2021clinically}.  Additionally, we expect that patients with larger drops in TIR week-over-week, and patients with higher mean glucose, are more likely to see a TIR benefit from clinician intervention.  Finally, it is clinically well-established that patients using insulin pumps tend to be better able to manage their TIR, and thus patients not using insulin pumps are more likely to see a TIR benefit from clinician intervention \cite{berget2019clinical}.

In Figure \ref{fig:interpret}, we show the values of the CATE predictions for the optimal action, as we vary patient features across panels.  
For each factor, the policy using clinician-informed representations matches exactly the clinical guidelines described in the previous paragraph: patients with lower TIR, a larger drop in TIR week-over-week, a higher mean glucose, or not using a pump will be prioritized for contact.  Conversely, the policies using black-box representations lack this interpretability and are badly misligned with clinical guidance: notably they prioritize patients with higher TIR, lower drops in TIR, and lower mean glucose.

\vspace{-1\baselineskip}
\section{Discussion} 
\vspace{-0.4\baselineskip}

We introduced an end-to-end pipeline for learning and evaluating policies induced by different CATE estimators across both black-box and low-dimensional, clinically-grounded state and action representations. Using data from clinical trials with remote patient monitoring for type 1 diabetes care, we learned clinician-informed policies that outperformed black-box-learned policies.

Iterating on the state and action representations could further improve policy performance. We expect that clustering-embedding-based action representations might produce more effective action representations with increased data, though interpretability would remain a challenge. Any algorithm would likely need a very large amount of data to learn meaningful state representations of the high-dimensional CGM traces that predict where actions fall in the high-dimensional message text space. With low-dimensional clinician-informed representations, it is much easier to learn the relationship between state and action representations (e.g. patients with more low CGM readings are more likely to get a message addressing low glucose). Future work could leverage larger datasets or synthetic data to understand how much data is necessary to learn useful embeddings for policy learning from high-dimensional clinical data.

Despite adjusting for all available information when estimating treatment effects and learning policies, unmeasured confounding may persist, which could bias our results. The only way to guarantee no unmeasured confounding would be to collect data where actions are taken randomly with known propensities, which might be infeasible in most healthcare settings. Future work could include sensitivity analyses to unmeasured confounding.

In our setting, the care team has capacity to take actions on K patients in each time period. Future work could examine a more general setting in which different actions have different capacity costs. Our analysis focuses on short-term outcomes; evaluating long-term effects requires additional assumptions or a randomized controlled trial of the learned policies over a longer period \citep{ferstad_smart_2024, collins_multiphase_2007}. 

Our approach can improve digital health interventions with large state and action spaces when clinical domain knowledge is available. For instance, it could enhance interventions based on wearable sensor data (e.g., smart watches measuring pulse and activity). Exactly how the approach is applied will depend on the setting and which clinician-informed state and action representations are available. Our approach enables evaluating policies learned from different candidate representations. Successful deployment of successful targeting policies could boost intervention efficacy and patient outcomes, promoting digital health adoption. However, it is crucial to ensure that patients {\em not} selected for treatment by a learned policy receive alternative or complementary interventions. Future work requires identifying and evaluating such interventions to ensure all patients receive appropriate care, as well as ensuring equitable access to such interventions \cite{prahalad_equitable_2024}.

\clearpage

\acks{
This work was supported in part by the NIH via the Stanford Diabetes Research Center (1P30DK11607401) and grant no. R18DK122422. Funding support was also received from the Helmsley Charitable Trust (G-2002-04251-2), National Science Foundation (2205084), the Stanford Institute for Human-Centered Artificial Intelligence (HAI), AFOSR Grant FA9550-21-1-0397, ONR Grant N00014-22-1-2110, the Stanford Data Science Scholars Program, and Stanford Maternal \& Child Health Research Institute. The Stanford REDCap platform (http://redcap.stanford.edu) is developed and operated by Stanford Medicine Research Technology team. The REDCap platform services at Stanford are subsidized by a) Stanford School of Medicine Research Office, and b) the National Center for Research Resources and the National Center for Advancing Translational Sciences, National Institutes of Health, through grant UL1 TR003142. Funding for devices and some CGM supplies was provided by a grant through the Lucile Packard Children’s Hospital Auxiliaries Endowment. EBF is a Chan Zuckerberg Biohub – San Francisco Investigator. The funders had no role in study design, data collection and analysis, decision to publish or preparation of the manuscript.
}

\bibliography{references}

\newpage

\appendix
\onecolumn  %

\section{Data summary}\label{appendix:data}

Table \ref{tab:summary_stats} presents the summary statistics of the patient population across the three data splits. We have more than a year of data and 10+ messages received (treatments) for most of the patients in our datasets. The data splits have similar summary statistics, as expected with random splitting.

\begin{table}[H]
\centering
\caption{Summary statistics by dataset/split}\label{tab:dataset_split}
\resizebox{\textwidth}{!}{
\begin{tabular}{lrrrr}\toprule
 &\multicolumn{3}{c}{\textbf{Dataset}} \\\cmidrule{2-4}
\textbf{Summary statistic} &Train &Validation &Test \\\midrule
N (\# of patients) &91 &95 &95 \\
Pilot study participants &26 &24 &25 \\
4T study participants &42 &42 &46 \\
TIPS study participants &24 &35 &27 \\
\# observations (days); mean (IQR) across patients &734 (408-1071) &691 (350-969) &699 (411-952) \\
\# messages received; mean (IQR) across patients &41 (17-60) &38 (12-58) &35 (14-52) \\
Age; mean (IQR) across patients &13 (10-17) &12 (9-16) &13 (10-17) \\
N patients using an insulin pump during study &66 &70 &63 \\
N patients using automated insulin delivery during study &51 &34 &41 \\
\bottomrule
\end{tabular}}
\label{tab:summary_stats}
\end{table}

\section{Computational resources}\label{appendix:compute}

To train nuisance models and CATE estimators with EconML and FLAML, we used an instance of Google Compute Engine machine type n2d-standard-224 with 224 vCPUs and 896 GB memory. We also did some training on a n2d-standard-64 instance (64 vCPUs, 256 GB RAM). Generating all of the results in the paper took less than a day with the n2d-standard-224.

To train the TS2Vec encoder, we used an instance of Google Compute Engine machine type n1-highmem-8 with 8 vCPUs, 52 GB memory, and one NVIDIA V100 GPU. Training the encoder took less than a day.

\newpage

\section{Action representations}\label{appendix:actionrep}

We generate clinical features with a few-shot prompt and language model (LM). Here's an example prompt and output.

\textbf{Prompt}
\vspace{-1.5\baselineskip}
\begin{lstlisting}[style=promptstyle]
You are an AI assistant who extracts structured JSON from messages sent by clinicians to patients with diabetes. 
* First identify whether the message recommends changing insulin (recommends_insulin_dose_change).
* If the message recommends an insulin change, label whether it is a basal / long acting insulin change (recommends_changing_basal_or_long_acting_insulin), a reminder to take correction doses (recommends_more_correction_doses), whether it adjusts the carb ratio at meal time (recommends_changing_carb_ratio), or if it has a reminder to bolus before meals (reminds_patient_to_bolus).
* Then identify the targets of the message, which can include nighttime glucose (recommends_insulin_change_at_night), daytime/mealtime glucose (recommends_insulin_change_during_the_day), high glucose or low time in range (recommendations_target_high_glucose_or_low_time_in_range), or low glucose (recommendations_target_low_glucose).
* Finally, determine if the message mentiones a recent visit (mentions_recent_visit), or knowledge of the patient's schedule (mentions_patient_schedule).

The output should be JSON with this structure: ... # see below

input: Looking at trends this past week compared to the previous week your average increased from 135 to 245. Has anything changed in your routine? Any insights into your daily routine would help me recommend dose changes. For now, I have decreased your bedtime correction target to 150. By giving correction insulin before bed we can help keep you in target range overnight. I would like to make further recommendation to help bring you down to target range more often , but again can you provide insights into your schedule ? How many times per day are you dosing insulin ? Are you giving insulin for both carbs and glucose level at mealtime?
output: {{
recommends_insulin_dose_change: True,
recommends_changing_basal_or_long_acting_insulin: False,
recommends_more_correction_doses:  True,
recommends_changing_carb_ratio:  False,
reminds_patient_to_bolus:  False,
recommends_insulin_change_at_night: True,
recommends_insulin_change_during_the_day: False,
recommendations_target_high_glucose_or_low_time_in_range:  True,
recommendations_target_low_glucose:  False,
mentions_recent_visit: False,
mentions_patient_schedule: False
}}
# + 4 more examples

input: X has an average blood glucose of 390 and is in target range (70-150) 1%
output:
\end{lstlisting}
\vspace{-1.5\baselineskip}

\textbf{Output}
\vspace{-1.5\baselineskip}
\begin{lstlisting}[style=outputstyle]
{
recommends_insulin_dose_change: True,
recommends_changing_basal_or_long_acting_insulin: True,
recommends_more_correction_doses: False,
recommends_changing_carb_ratio: False,
reminds_patient_to_bolus: False,
recommends_insulin_change_at_night: True,
recommends_insulin_change_during_the_day: False,
recommendations_target_high_glucose_or_low_time_in_range: True,
recommendations_target_low_glucose: False,
mentions_recent_visit: False,
mentions_patient_schedule: False
}
\end{lstlisting}

\begin{table}[H]\centering
\caption{Summary statistics across extracted clinical features}\label{tab:clinical_features}
\scriptsize
\begin{tabular}{lrr}\toprule
\textbf{Clinical feature} & \textbf{Total messages (\%)} \\\midrule
recommends\_insulin\_dose\_change &4,523 (18\%) \\
recommends\_changing\_basal\_or\_long\_acting\_insulin &2,606 (10\%) \\
recommends\_more\_correction\_doses &1,037 (4\%) \\
recommends\_changing\_carb\_ratio &1,795 (7\%) \\
reminds\_patient\_to\_bolus &1,167 (5\%) \\
recommends\_insulin\_change\_at\_night &2,802 (11\%) \\
recommends\_insulin\_change\_during\_the\_day &2,280 (9\%) \\
recommendations\_target\_high\_glucose\_or\_low\_time\_in\_range &4,183 (17\%) \\
recommendations\_target\_low\_glucose &3,861 (15\%) \\
mentions\_recent\_visit &1,510 (6\%) \\
mentions\_patient\_schedule &1,521 (6\%) \\
\bottomrule
\end{tabular}
\end{table}

\textbf{Defining discrete clinician-informed action representations.} To get discrete representations from the clinical features, we first estimate the average treatment effect on the treated to make sure we include features associated with much greater treatment effects than just receiving any message. Then we group the features into discrete representations with clinically different meanings:
\begin{itemize}
    \item Message treating highs and lows: \\ 
    \texttt{(recommendations\_target\_low\_glucose) AND \\(recommendations\_target\_high\_glucose\_or\_low\_time\_in\_range OR \\
    recommends\_more\_correction\_doses OR reminds\_patient\_to\_bolus)}
    \item Message treating highs only: \\
    \texttt{(NOT recommendations\_target\_low\_glucose) AND \\(recommendations\_target\_high\_glucose\_or\_low\_time\_in\_range \\
    OR recommends\_more\_correction\_doses OR reminds\_patient\_to\_bolus)}
    \item Message treating lows only: \\
    \texttt{(recommendations\_target\_low\_glucose) AND NOT \\(recommendations\_target\_high\_glucose\_or\_low\_time\_in\_range \\
    OR recommends\_more\_correction\_doses OR reminds\_patient\_to\_bolus)}
    \item Other Message: Messages falling into none of the categories above.
\end{itemize}

\textbf{Baseline representations.} To extract features from the raw messages, we generate 728-dimensional text embeddings with PaLM 2 \citep{anil_palm_2023}. Then we learn a clustering of those embeddings using K-means on the training data with K=4 to match the cardinality of the clinician-informed action representation. Messages in the other datasets are mapped to the closest cluster centroid.

\newpage
\section{State representations}\label{appendix:staterep}

\subsection{Clinical features and representations}

\noindent Full list of clinical state features:
\vspace{-\baselineskip}
\begin{multicols}{2}
\scriptsize
\texttt{g\_7dr}: Mean glucose last 7 days \\
\texttt{very\_low\_7dr}: Prop. CGM readings $<$ 54 mg/dL last 7 days \\
\texttt{low\_7dr}: Prop. CGM readings $<$ 70 mg/dL last 7 days \\
\texttt{in\_range\_7dr}: Prop. CGM readings 70-180 mg/dL last 7 days \\
\texttt{high\_7dr}: Prop. CGM readings $>$ 180 mg/dL last 7 days \\
\texttt{very\_high\_7dr}: Prop. CGM readings $>$ 250 mg/dL last 7 days  \\
\texttt{gri\_7dr}: Glycemia Risk Index \citep{klonoff_glycemia_2023} last 7 days \\
\texttt{g\_14dr}: Mean glucose last 14 days \\
\texttt{very\_low\_14dr}: Prop. CGM readings $<$ 54 mg/dL last 14 days \\
\texttt{low\_14dr}: Prop. CGM readings $<$ 70 mg/dL last 14 days \\
\texttt{in\_range\_14dr}: Prop. CGM readings 70-180 mg/dL last 14 days \\
\texttt{high\_14dr}: Prop. CGM readings $>$ 180 mg/dL last 14 days \\
\texttt{very\_high\_14dr}: Prop. CGM readings $>$ 250 mg/dL last 14 days \\
\texttt{gri\_14dr}: Glycemia Risk Index \citep{klonoff_glycemia_2023} last 14 days \\
\texttt{night\_very\_low\_7dr}: Prop. CGM readings $<$ 54 mg/dL last 7 days at night time (11pm-5am) \\
\texttt{night\_low\_7dr}: Prop. CGM readings $<$ 70 mg/dL last 7 days at night time (11pm-5am) \\
\texttt{night\_high\_7dr}: Prop. CGM readings $>$ 180 mg/dL last 7 days at night time (11pm-5am) \\
\texttt{night\_very\_high\_7dr}: Prop. CGM readings $>$ 250 mg/dL last 7 days at night time (11pm-5am) \\
\texttt{day\_very\_low\_7dr}: Prop. CGM readings $<$ 54 mg/dL last 7 days at day time (6am-10pm) \\
\texttt{day\_low\_7dr}: Prop. CGM readings $<$ 70 mg/dL last 7 days at day time (6am-10pm) \\
\texttt{day\_high\_7dr}: Prop. CGM readings $>$ 180 mg/dL last 7 days at day time (6am-10pm) \\
\texttt{day\_very\_high\_7dr}: Prop. CGM readings $>$ 250 mg/dL last 7 days at day time (6am-10pm) \\
\texttt{time\_worn\_7dr}: Prop. of time with CGM readings last 7 days \\
\texttt{night\_worn\_7dr}: Prop. of time with CGM readings last 7 days at night time (11pm-5am) \\
\texttt{day\_worn\_7dr}: Prop. of time with CGM readings last 7 days at day time (6am-10pm) \\
\texttt{gri\_7dr\_7d\_delta}: Difference in \texttt{gri\_7d} between today and 7 days ago \\
\texttt{very\_low\_7dr\_7d\_delta}: Difference in \texttt{very\_low\_7dr} between today and 7 days ago \\
\texttt{low\_7dr\_7d\_delta}: Difference in \texttt{low\_7dr} between today and 7 days ago \\
\texttt{in\_range\_7dr\_7d\_delta}: Difference in \texttt{in\_range\_7dr} between today and 7 days ago \\
\texttt{very\_high\_7dr\_7d\_delta}: Difference in \texttt{very\_high\_7dr} between today and 7 days ago \\
\texttt{night\_very\_low\_7dr\_7d\_delta}: Difference in \texttt{night\_very\_low\_7dr} between today and 7 days ago  \\
\texttt{night\_low\_7dr\_7d\_delta}: Difference in \texttt{night\_low\_7dr} between today and 7 days ago \\
\texttt{night\_high\_7dr\_7d\_delta}: Difference in \texttt{night\_high\_7dr} between today and 7 days ago \\
\texttt{sexF}: Indicator equal to 1 for female patients, 0 otherwise. \\
\texttt{public\_insurance}: Indicator equal to 1 for publicly insured patients, 0 otherwise \\
\texttt{english\_primary\_language}: Indicator equal to 1 for patients with English as their preferred language, 0 otherwise \\
\texttt{pop\_pilot}: Indicator equal to 1 for patients enrolled in the 4T Pilot study \citep{prahalad_teamwork_2022}, 0 otherwise \\
\texttt{pop\_4T\_1}: Indicator equal to 1 for patients enrolled in the 4T  Study 1 \citep{prahalad_equitable_2024}, 0 otherwise \\
\texttt{pop\_TIPS}: Indicator equal to 1 for patients enrolled in the TIPS  Study \citep{scheinker_new_2022}, 0 otherwise \\
\texttt{age}: Age of patient \\
\texttt{months\_since\_onset}: Months since onset of type 1 diabetes \\
\texttt{using\_pump}: Indicator equal to 1 for patients using an insulin pump, 0 otherwise \\
\texttt{using\_aid}: Indicator equal to 1 for patients using automated insulin delivery (closed loop), 0 otherwise \\
\texttt{days\_since\_msg}: Days since the last time the patient received a message \\
\texttt{large\_tir\_drop}: Indicator equal to 1 for patients with \texttt{in\_range\_7dr\_7d\_delta<-0.15}, 0 otherwise (used by clinicians for risk stratification) \\
\texttt{low\_tir}: Indicator equal to 1 for patients with \texttt{in\_range\_7dr<0.65}, 0 otherwise (used by clinicians for risk stratification) \\
\texttt{lows}: Indicator equal to 1 for patients with \texttt{lows\_7dr>0.04}, 0 otherwise (used by clinicians for risk stratification \\
\texttt{very\_lows}: Indicator equal to 1 for patients with \texttt{very\_lows\_7dr>0.01}, 0 otherwise (used by clinicians for risk stratification \\
\end{multicols}

We create three different representations using these features:
\begin{enumerate}
    \item Full: includes all features listed above
    \item Subset selected with ML: Top features based on XGBoost variable importance (SHAP) scores when training a model to predicting observed rewards in the training data. Includes \texttt{large\_tir\_drop, in\_range\_7dr\_7d\_delta, in\_range\_14dr, in\_range\_7dr, low\_7dr, using\_pump, using\_aid, time\_worn\_7dr, day\_worn\_7dr, day\_low\_7dr, night\_high\_7dr\_7d\_delta, g\_7dr, months\_since\_onset, gri\_7dr\_7d\_delta}
    \item TIDE subset selected by clinicians: Features shown to clinicians when they review patients in TIDE. Includes \texttt{very\_low\_7dr, low\_7dr, in\_range\_7dr, g\_7dr, using\_pump, in\_range\_7dr\_7d\_delta, large\_tir\_drop, low\_tir, lows, very\_lows, pop\_4T\_1, pop\_4T\_2, pop\_TIPS} 
\end{enumerate}

\newpage

\subsection{Baseline representations}

\textbf{TS2Vec.} We generate low-dimensional representations of the CGM traces using TS2Vec \citep{yue_ts2vec_2021}. We train the encoder on the training data, feeding in two weeks of CGM readings (4,032-dimensional vectors). We use the default hyperparameters in the code at https://github.com/zhihanyue/ts2vec/, with n\_epochs=1 and output\_dims=8. We also tested training for 2 and 3 epochs and found very similar results to n\_epochs=1 when using the representations to fit CATE estimators.

\textbf{UMAP.} We project two weeks of CGM readings into low-dimensional representations using UMAP as implemented at https://github.com/lmcinnes/umap. We tested projecting down to between 1 and 10 dimensions and then picked the number of dimensions that performed best at predicting observed rewards in the training data. We ended up picking n\_components=4 for the projections used for CATE estimation.

\section{Additional policy evaluation results}\label{appendix:additional_results}

\begin{figure}[H]
\centering 
\input{figures/toc.tex}
\caption{Illustrative TOC curve.}
\label{fig:toc} 
\end{figure}
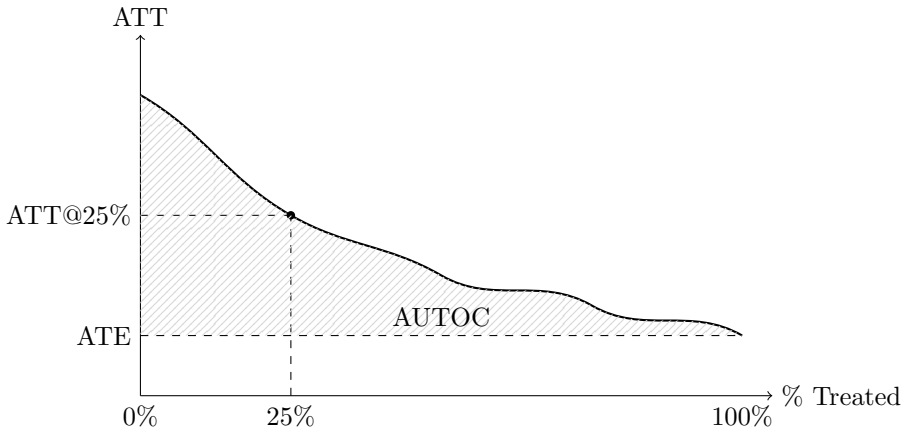

In plotting our evaluation results for a given policy, we visualize the {\em targeting operator characteristic (TOC) curve} \citep{sverdrup_qini_2023, yadlowsky_evaluating_2021}: this is a plot of $\widehat{\mathsf{ATT}}_K(\pi)$ against $K/N$; when $K = N$, we obtain the average treatment effect (ATE) of the policy.  We generate confidence intervals along the TOC by bootstrapping patients in the evaluation data. A common evaluation metric used to evaluate targeting policies is the Area Under the TOC curve (AUTOC), which is the area between the ATT and ATE integrated over K from 0 to 1. See Figure \ref{fig:toc} for reference. In this section, we report the ATT@25\% and AUTOC values for each state and action representation across  CATE estimators.

\begin{figure}[H]
\centering 
\includegraphics[width=1\linewidth]{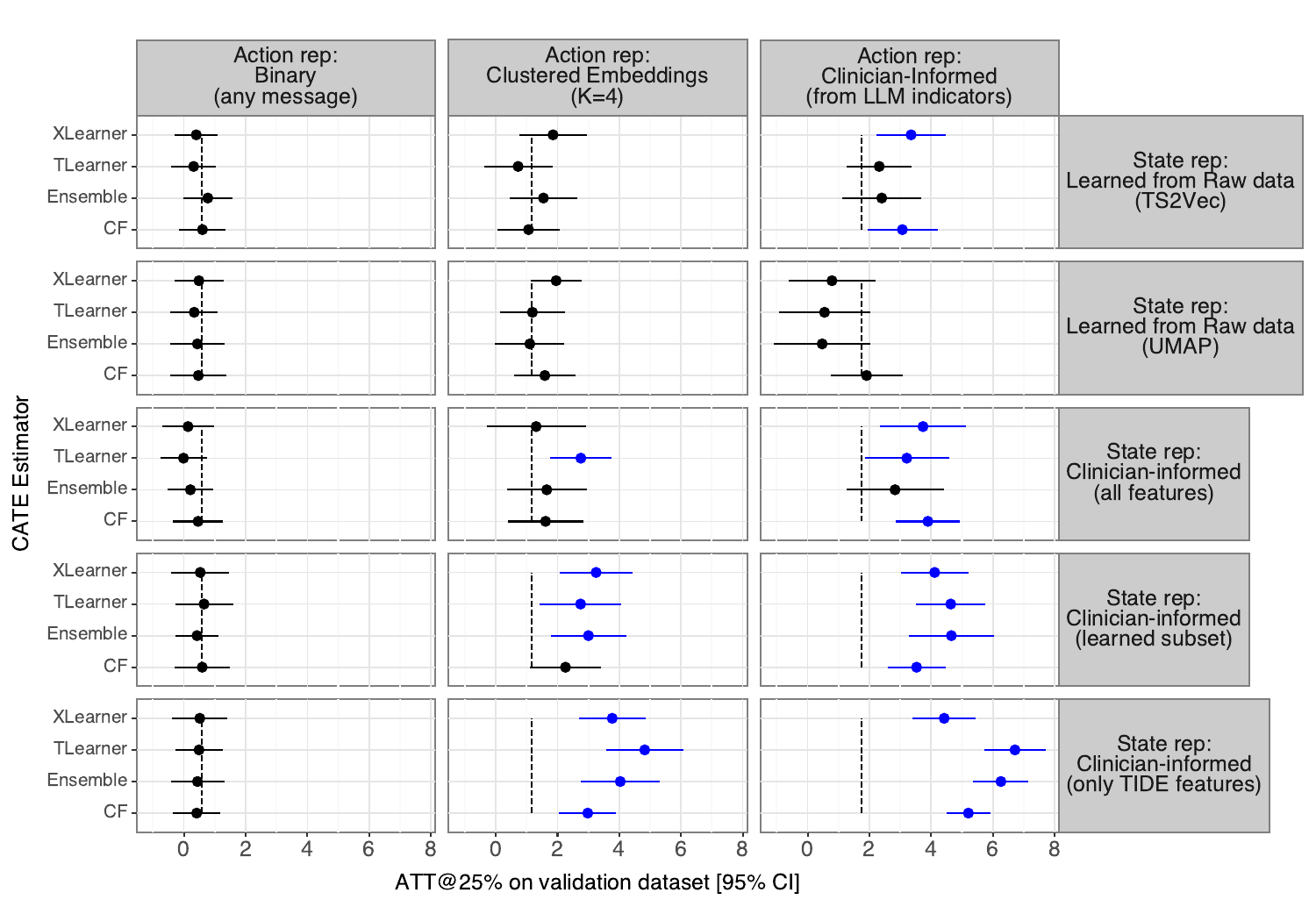} 
\caption{ATT@25\% for each state (rows) and action (columns) representation across CATE estimators. Results with the S-Learner and DR Forest CATE estimators are excluded because they never outperformed random targeting.}
\label{fig:att25_all_cates} 
\end{figure} 

\begin{figure}[H]
\centering 
\includegraphics[width=1\linewidth]{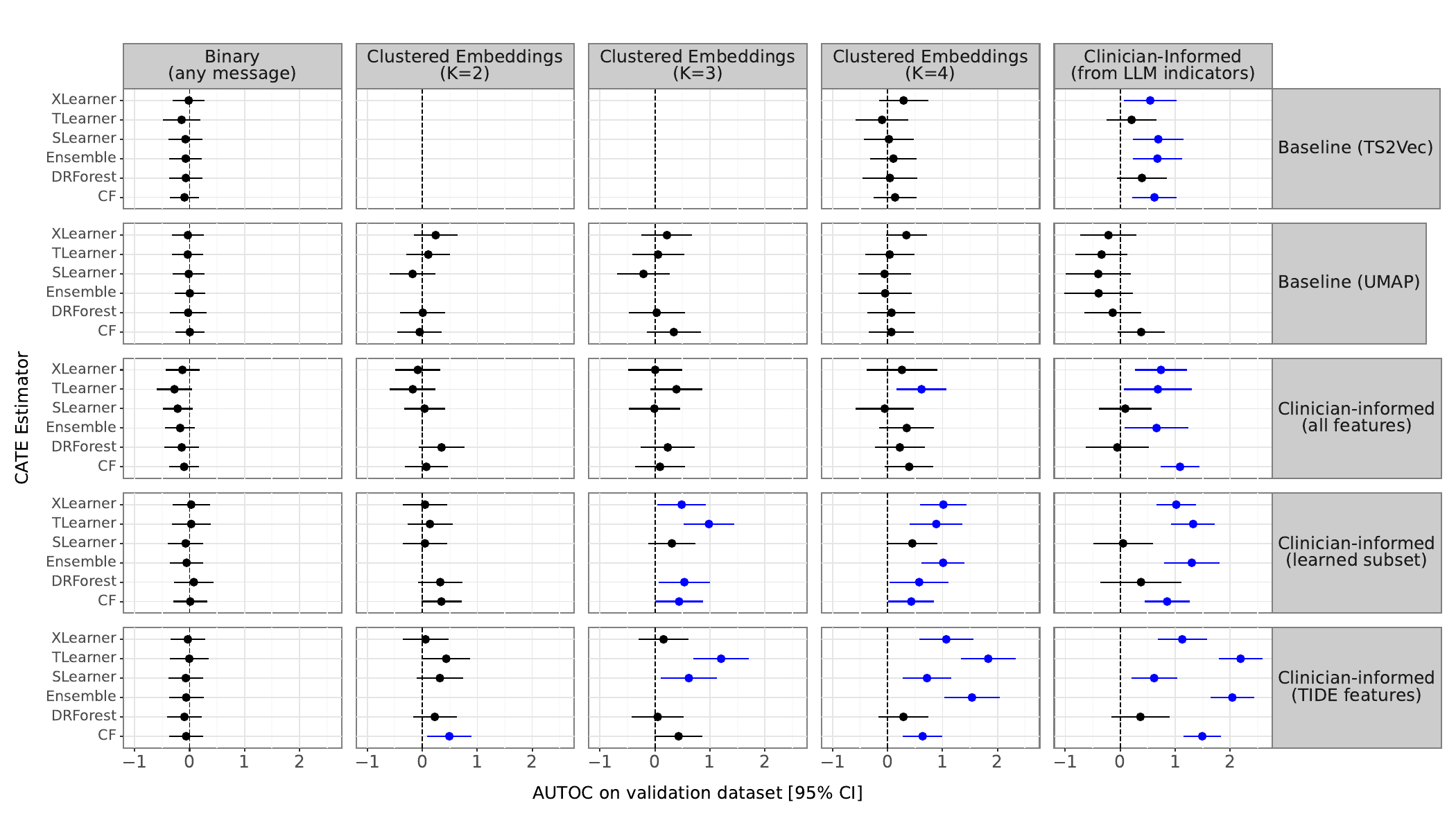} 
\caption{AUTOCs for each state (rows) and action (columns) representation across CATE estimators}
\label{fig:autocs} 
\end{figure} 

\begin{figure}[H]
\centering 
\includegraphics[width=0.67\linewidth]{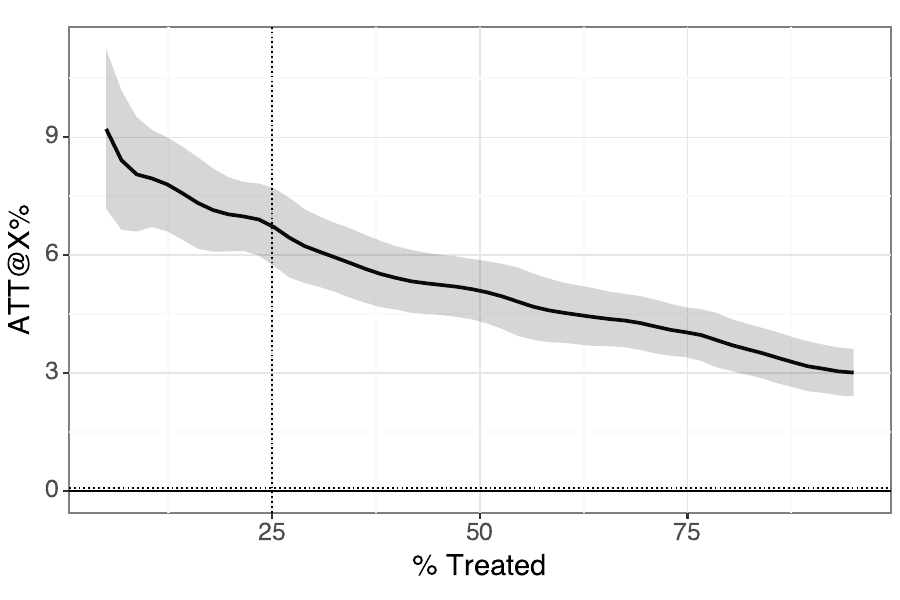} 
\caption{Example TOC curve for the top performing policy on the validation set:\\ State representation = Clinician-informed (TIDE)\\ 
Action representation = Clinician-informed \\
CATE estimator = TLearner.}
\label{fig:ex_toc} 
\end{figure}

\newpage
\section{Sensitivity analysis to adding longer patient history in control covariates when evaluating policies}\label{appendix:prop_sensitivity}

\begin{figure}[H]
\centering 
\includegraphics[width=1\linewidth]{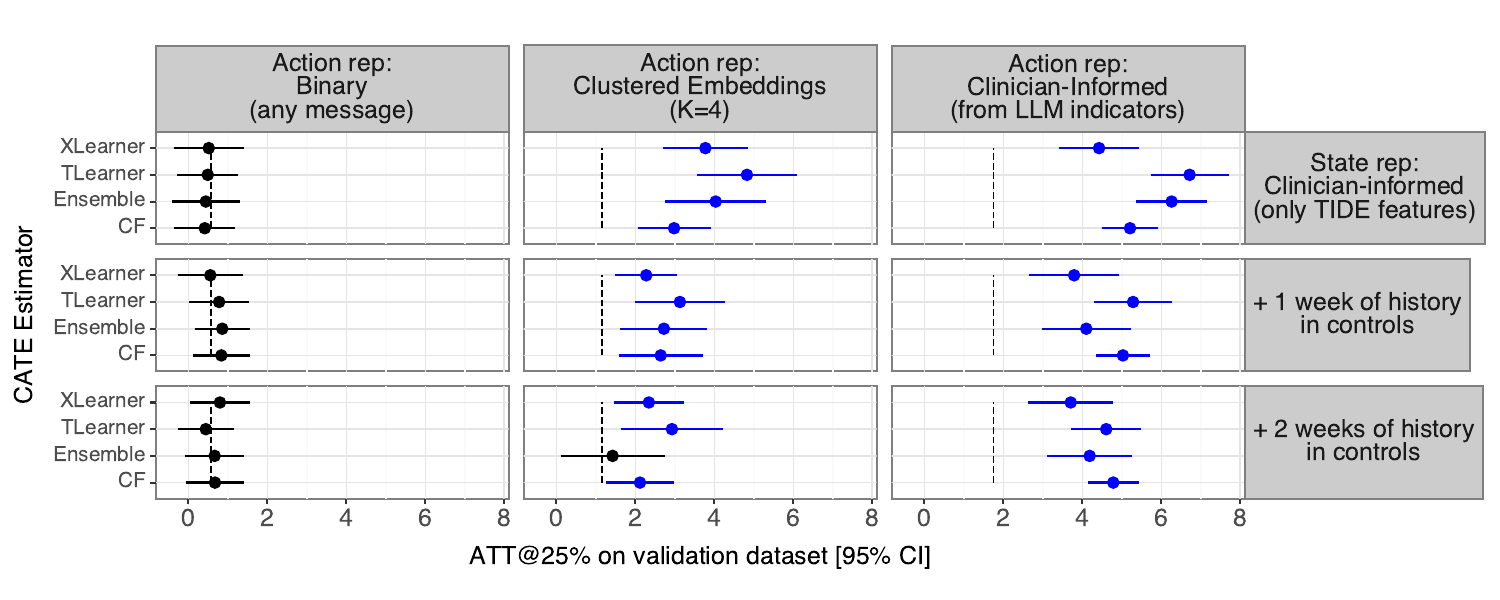} 
\caption{ATT@25\% when including more history in the control covariates.}
\label{fig:robustness_covariate_eval} 
\end{figure}

In order to understand if we are getting biased policy evaluation results by only using two weeks of CGM data to construct our control covariates, we also evaluated the best-performing state representation with control covariates that include longer histories (up to 4 weeks), including indicators for messages in previous weeks. The ATT@25\% when additional weeks of history are included in control covariates are shown in Figure \ref{fig:robustness_covariate_eval}. We see that ATT@25\% remains much higher than the ATEs (dashed vertical lines) when we are using clinician-informed action representations (right-most column). There is a slight drop-off in estimated performance as we add  history to the control covariates though. This could be noise, but future work might include more sensitivity analyses, and also adding more weeks of history to the state representations.

\newpage
\section{Additional results and proofs}\label{appendix:proofs}

\begin{proposition}
\label{prop:pi_star}
Fix $K$, and let $\pi^* = \pi(\cdot | \tau, K)$.  Let $\pi$ be any other policy that targets at most $K$ patients (i.e., selects at most $K$ patients to receive a non-control action).  Then $\mathsf{ATT}_K(\pi^*) \geq \mathsf{ATT}_K(\pi)$.
\end{proposition}

{\em Proof of Proposition \ref{prop:pi_star}.}
Fix a state vector $(s_1, \ldots, s_N)$ (sampled i.i.d.~from the superpopulation).  For each patient $i$, let $a_i^* = \arg \max_{\tilde{a} \in \mathcal{A}} \tau(s_i, \tilde{a})$, i.e., the optimal action for that patient.  Let $\alpha_i = \tau(s_i, a_i^*)$, and let $(\alpha_{(1)}, \ldots, \alpha_{(N)})$ be the same values sorted in descending order.  Observe that since $\pi^*$ selects the $K$ highest ranked patients according to $\alpha_i$, it follows that for any subset $I_K \subset \{ 1, \ldots, N\}$ of cardinality $K$, there holds:
\[ \sum_{i = 1}^N \tau(s_i, \pi_i^*(s)) = \sum_{i = 1}^K \alpha_{(i)} \geq \sum_{i \in I_K} \alpha_i. \]
Now let $I_K$ be the set of (at most $K$) patients who receive non-control actions under another feasible policy $\pi$.  The preceding allows us to conclude that:
\[ \sum_{i = 1}^N \tau(s_i, \pi_i^*(s)) \geq \sum_{i = 1}^N \tau(s_i, \pi_i(s)),\]
since $\alpha_i \geq \tau(s_i, \pi_i(s))$ for all $i$.  Dividing by $K$ and taking expectations over $s$ concludes the proof.\hfill$\Box$\\

{\em Proof of Theorem \ref{thm:optimality}}.  
First, we use the Skorohod representation theorem to construct a probability space on which $\sup_{s \in \mathcal{S},a \in \mathcal{A}} | \tau(s,a) - \hat{\tau}_N(s,a) | \to 0$ almost surely.  Next, suppose that $s = (s_1, \ldots, s_N)$ is an i.i.d.~sample of $N$ patients from the superpopulation.  Note that $\pi_N$ applies optimal treatments to the top $K$ ranked patients according to $\hat{\tau}_N$.  Therefore:
\[ \frac{1}{K_N} \sum_{i = 1}^N \hat{\tau}_N(s_i, \pi_{N,i}(s)) \geq  \frac{1}{K_N} \sum_{i = 1}^N \hat{\tau}_N(s_i, \pi_{N,i}^*(s)).\]
For each $i$ and $N$, define $\Delta_{i,N} = \hat{\tau}_N(s_i, \pi_{N,i}(s)) - \tau(s_i, \pi_{N,i}(s))$, and define $\Delta^*_{i,N} = \hat{\tau}_N(s_i, \pi_{N,i}^*(s)) - \tau(s_i, \pi_{N,i}^*(s))$.  Then the preceding inequality becomes:
\[  \frac{1}{K_N} \sum_{i = 1}^N ( \tau(s_i, \pi_{N,i}(s)) - \tau(s_i, \pi_{N,i}^*(s)) ) + \frac{1}{K_N} \sum_{i = 1}^N  ( \Delta_{i,N} - \Delta^*_{i,N}) \geq 0. \]
Note the second summation on the left hand side converges to zero almost surely.  On the other hand, because $\pi_N^*$ is the oracle optimal policy, we also have:
\[ \frac{1}{K_N} \sum_{i = 1}^N ( \tau(s_i, \pi_{N,i}(s)) - \tau(s_i, \pi_{N,i}^*(s)) ) \leq 0.\]
Therefore we conclude:
\[ \lim_{N \to \infty} \frac{1}{K_N} \sum_{i = 1}^N ( \tau(s_i, \pi_{N,i}(s)) - \tau(s_i, \pi_{N,i}^*(s)) ) = 0\]
almost surely.

Since treatment effects are bounded, we can apply the bounded convergence theorem to conclude that $\mathsf{ATT}_{K_N}(\pi_N) - \mathsf{ATT}_{K_N}(\pi_N^*) \to 0$, as required.
\hfill$\Box$\\

\end{document}

%% file: figures/toc.tex
\usetikzlibrary{patterns, calc, decorations.pathreplacing}

\begin{tikzpicture}[scale=0.8]
    \coordinate (A) at (0, 5);
    \coordinate (B) at (2.5, 3);
    \coordinate (C) at (5, 2);
    \coordinate (D) at (7.5, 1.5);
    \coordinate (E) at (10, 1);

    \draw[->] (0, 0) -- (10.5, 0) node[right] {$\%$ Treated};
    \draw[->] (0, 0) -- (0, 6) node[above] {ATT};

    \draw[thick] (A) to[out=-30, in=150] (B) to[out=-30, in=150] (C) to[out=-30, in=150] (D) to[out=-30, in=150] (E);

    \node[below] at (0, 0) {$0\%$};
    \node[below] at (10,0) {$100\%$};
    \node[left] at (0,1) {ATE};

    \coordinate (X) at (2.5, 0);
    \draw[dashed] (X) -- ($(X)!(B)!(X)$);
    \node[below] at (X) {$25\%$};
    \draw[dashed] ($(X)!(B)!(X)$) -- (B) -- (0, 3);
    \node[left] at (0, 3) {ATT@$25\%$};
    \fill (B) circle (2pt);

    \fill[pattern=north east lines, pattern color=gray!30] (A) to[out=-30, in=150] (B) to[out=-30, in=150] (C) to[out=-30, in=150] (D) to[out=-30, in=150] (E) -- (E-|A) -- cycle;
    \node[above] at (5,1) {AUTOC};

    \draw[dashed] (0, 1) -- (E);
    
\end{tikzpicture}